\documentclass[10pt,twocolumn,letterpaper]{article}

\usepackage{iccv}
\usepackage{times}
\usepackage{epsfig}
\usepackage{graphicx}
\usepackage{amsmath}
\usepackage{amssymb}

\usepackage{listings}
\usepackage{xcolor}
\usepackage{bbm}            
\usepackage{booktabs}       
\usepackage{multirow}       
\usepackage[breaklinks=true,bookmarks=false]{hyperref}

\iccvfinalcopy 

\def\iccvPaperID{****} 

\ificcvfinal\fi

\begin{document}

\title{Go Small and Similar: A Simple Output Decay Brings Better Performance}

\author{Xuan Cheng\thanks{University of Electronic Science and Technology of China}\\
{\tt\small cs\_xuancheng@std.uestc.edu.cn}
\and
Tianshu Xie$^{*}$\thanks{Contributed equally.}\\
{\tt\small tianshuxie@std.uestc.edu.cn}
\and
Xiaomin Wang$^{*}$\\
{\tt\small xmwang@uestc.edu.cn}
\and
Jiali Deng$^{*}$ \\
{\tt\small julia\_d@163.com}
\and
Minghui Liu$^{*}$ \\
{\tt\small minghuiliuuestc@163.com}
\and
Ming Liu$^{*}$ \thanks{Corresponding author}\\
{\tt\small csmliu@uestc.edu.cn}
}

\maketitle
\ificcvfinal\thispagestyle{empty}\fi

\begin{abstract}
Regularization and data augmentation methods have been widely used and become increasingly indispensable in deep learning training. Researchers who devote themselves to this have considered various possibilities. But so far, there has been little discussion about regularizing outputs of the model. This paper begins with empirical observations that better performances are significantly associated with output distributions, that have smaller average values and variances. By audaciously assuming there is causality involved, we propose a novel regularization term, called Output Decay, that enforces the model to assign smaller and similar output values on each class. Though being counter-intuitive, such a small modification result in a remarkable improvement on performance. Extensive experiments demonstrate the wide applicability, versatility, and compatibility of Output Decay.
\end{abstract}

\section{Introduction}
Improving the performance of deep convolutional neural networks (CNNs) has long been a question of great interest in a wide range of fields. Recent improvements in network structure and computer hardware have indeed brought promising performances, but they have also entailed increased risks of overfitting. To combat with this issue, a considerable literature has grown up around the themes of data augmentation and regularization. Taken together, these techniques have focused on various details of training, ranging from the inputs (Data Augmentations\cite{yun2019cutmix, devries2017improved, cubuk2018autoaugment, lim2019fast, zhong2020random}), to labels(Label Smoothing~\cite{szegedy2016rethinking}), to parameters (Weight Decay~\cite{krogh1991a}), to depths (Stochastic Depths~\cite{huang2016deep}), to neurons (Dropout~\cite{srivastava2014dropout}), to feature maps (Batch Normalization~\cite{ioffe2015batch}, DropBlock~\cite{ghiasi2018dropblock}) and so on. However, far little attention has been paid to regularizing the output distribution of the network.

The outputs of the network can be viewed as the projections of higher-dimensional features of images, namely, the output distribution itself implies the network’s perception of images. Actually, researchers have shown an increased interest in utilizing the knowledge embedded in the output distribution. Knowledge distillation~\cite{hinton2015distilling} believes the probabilities assigned to each class are part of the knowledge and exploit this by forcing a small network to assign the same probabilities as a large network that generalizes well. When estimating confidence of network, temperature scaling~\cite{guo2017on} that reduces all logits by a temperature factor is quite effective at calibrating a model’s predictions. So it can be inferred that regularizing the output distribution should also be able to play an important role in improving the performance of the network and addressing overfitting.

To motivate our regularizer, we perform several experiments to explore whether there is a connection between the output and performance of the model. Results empirically suggest that as performances of the models increase, the mean values and variances of outputs are smaller. Consequently, we suppose it is possible to further improve the model’s performance by shrinking the mean values and variances of the output distributions. In this work, we consider achieving this aim by introducing a complement loss Output Decay. This regularization term encourages the model to produce equal output value for each class. Experiments show that OD\footnote{Throughout this paper, the term 'OD' will refer to Output Decay.} achieves consistent performance gains on various benchmark datasets using multiple network architectures. Due to its inherent simplicity, this regularization term enjoys a plug-and-play nature, \ie it can be easily assembled into any programs without worrying about it having conflict with original parts. It is hoped that this research will contribute to a deeper understanding of the effect of outputs.

To offers some important insights about what does model learn with Output Decay, we explore several intriguing properties of models trained with OD. We begin by exploring whether OD could make the outputs smaller. By plotting reality diagram and computing ECE, we show that OD can effectively remedy the miscalibration phenomenon of networks. Then we find the effect of OD on parameters mainly happens on the fully-connected layer. By diminishing the parameters, OD helps raise the model's label noise robustness. The experiments in Section~\ref{practical}.A prove this. It will then go on to the feature activation parts, where we observe a very sparse activation situation, compared with vanilla training. This represents the network would have better feature selection capability and some overfitting tolerance. This is also evidenced by empirical experiments in Section~\ref{practical}.B. Lastly, we visualize and compare the feature distributions learned by models with and without OD. We find that OD leads to better and more discriminative feature learning.

\subsection{Summary of Contribution}
\begin{itemize}
	\item We provide observations that there are negative associations between the performance of the model and the outputs, namely, the better the network performance, the smaller the mean and variance of the output. (Section~\ref{motivation})
	\item Motivated by these observations, we propose a direct way to regularize the outputs, called Output Decay. It’s very simple and requires even one line of code modification. (Section~\ref{method})
	\item We explore and describe several intriguing properties regarding how and why OD brings better performance. (Section~\ref{visualize})
	\item We demonstrate the wide applicability and compatibility of our method. Specifically, we experiment with different architectures, different benchmark datasets, different combinations with other techniques, and some practical scenarios including label noise problem and small sample size problem. (Section~\ref{experiments})
\end{itemize}
\section{Observations}
\label{motivation}
Before we delve deeper into the detailed design of our method, let us first consider a question: \textit{is there any association between the outputs and the performance of the network? Or as commonly stated, what kind of outputs should be considered good for the model}?

To answer the question, we investigated two factors that might relate to the performance of the network: the magnitude and the standard deviation of the outputs. Specifically, we used the pretrained publicly available models provided by \textit{Pytorch}~\cite{paszke2019pytorch}. To eliminate interference of different network structures on outputs, we chose the Vgg~\cite{simonyan2014very} architecture for comparison (See the Appendix~\ref{more_observation} for similar analytic results of ResNet and DenseNet architectures). Due to the fact that most pretrained models have suffered from overfitting on the training samples, we evaluated the models’ outputs on the test samples of ImageNet~\cite{russakovsky2015imagenet}. We added the outputs all together to represent the magnitude of the outputs. As for the standard deviation of the outputs, we first calculated the standard deviation of each sample and then averaged them.

Figure~\ref{motivation-curve} depicts how the sum and the mean standard deviation of the outputs changes as the accuracy of the models increases. Surprisingly, with increasing accuracy, the magnitude of the models’ outputs shows a significant downward trend ($R^2=0.983$, **$p=0.008<.01$). The same strong and negative association exists between the mean standard deviation of the outputs and the performance of models ($R^2=0.956$, *$p=0.022<.05$). 

\begin{figure}[h]
	\centering
	\includegraphics[width=0.99\columnwidth]{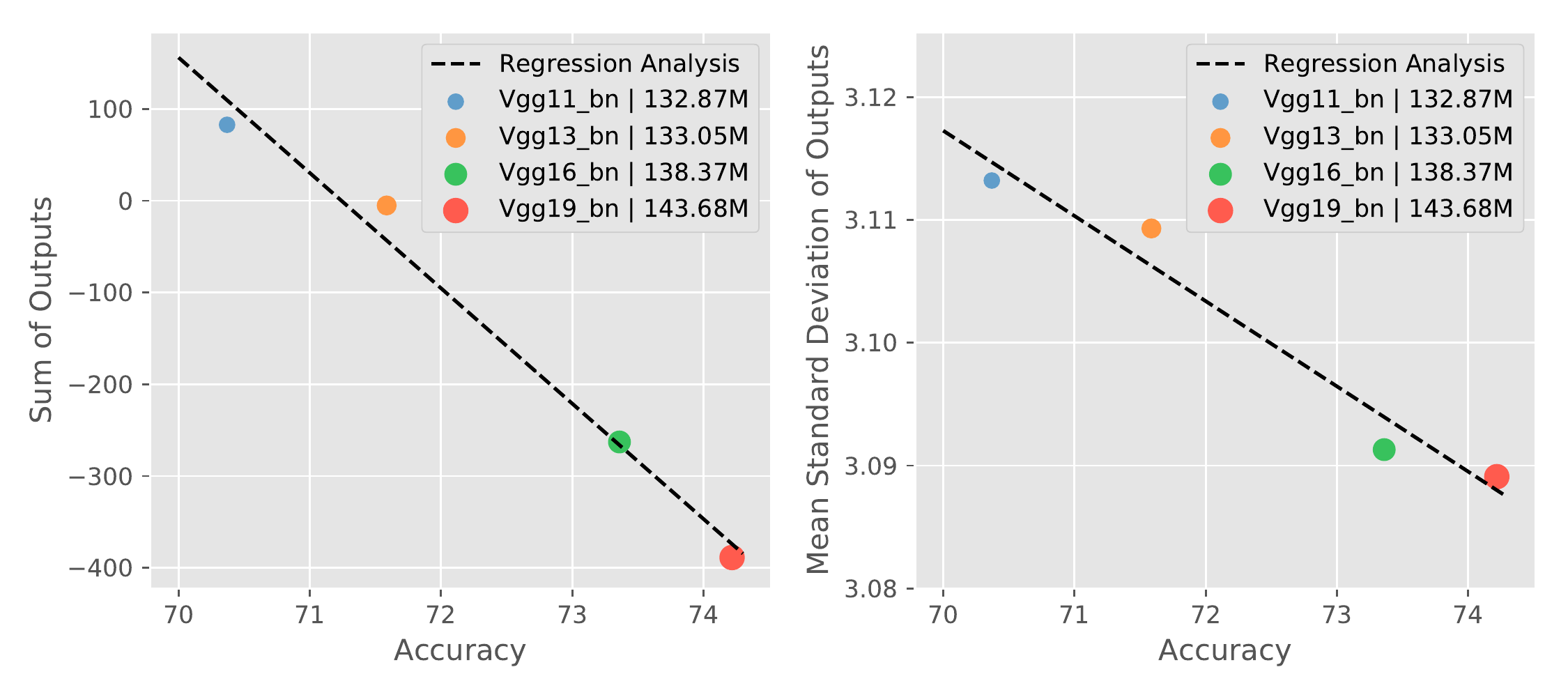}
	\caption{Connections between the models' performance and magnitudes(left)/variations(right) of the model's outputs. Node size represents size of parameters. The dashed lines denote the estimated results of regression analysis. }
	\label{motivation-curve}
\end{figure}

These results suggest that, the more powerful the model is, the more regularized its outputs become. This brings us the following questions: \textit{is it possible that the regularized outputs lead to the improvements of performance to some extent? If so, how can we utilize this in a more effective and easy-to-use way?} We will see how to address this issue next.
\section{Method}
\label{method}
Based on the observations and assumptions mentioned above, two properties of the outputs need to be regularized. That is to say, we should not only make outputs smaller, but also make outputs more uniformly distributed. Consequently, we propose to add an extra term to the cost function, a term called Output Decay. It is surprisingly simple. Consider $x\in R^{B\times C\times W\times H}$ and $f(x)\in R^{B\times N}$ denote a mini-batch of training images and their corresponding unnormalized outputs. $B$ and $N$ represent the mini-batch size and the number of classes in the dataset. Then, the OD term would be:
\begin{equation}
\mathcal{L}_{od} = \frac{1}{2} \cdot \frac{1}{B\times N}\cdot \sum_{i=1}^{B}\sum_{j=1}^{N}{(f(x)_{i,j}-c)}^2
\end{equation}
where $c$ denotes the decay level. Then the regularized cross-entropy would be:
\begin{equation}
\mathcal{L} = \mathcal{L}_{ce} +\beta \mathcal{L}_{od}
\end{equation}
where $\mathcal{L}_{ce}$ represents the original cross-entropy function and $\beta$ denotes the decay coefficient. 

Intuitively, the goal of our regularization term is to make each component of the outputs get closer to the decay level $c$. By presetting a small value of $c$, it not only ensures the outputs to become smaller but also guarantees the output distribution to be as much uniformly distributed as possible. This is the key of Output Decay and what we meant by \textit{go small and similar} in the title. Having said that, large outputs can also be allowed if they considerably improve the original loss function that corresponds to the classification tasks. Put another way, our OD term can be viewed as a way of compromising between regularizing the output distribution and minimizing the original loss function. A scalar $\beta$ is used for balancing these two terms, when $\beta$ is small we prefer to minimize the original loss function, but when $\beta$ is large we push harder on regularizing outputs distribution. It may be hard to set both the decay coefficient $\beta$ and the decay level $c$ without expert knowledge on the task. We can mitigate this issue easily by treating them as hyper-parameters so that we can exhaustively evaluate the accuracy for the predefined hyper-parameter candidates with a validation dataset, which can be performed in parallel.

Output Decay is simple and incurs a negligible computational overhead. This plug-and-play regularization term can be conveniently incorporated into almost any existing deep learning projects by merely adding one line of code after evaluating the original loss function. A minimal working example with a mini-batch in PyTorch is demonstrated below to show the additional one line of code:
\lstset{
	basicstyle          =   \ttfamily,          
	keywordstyle        =   \ttfamily,          
	commentstyle        =   \color{teal}\ttfamily,  
	stringstyle         =   \ttfamily,  
	flexiblecolumns,                
	showspaces          = false,  
	showstringspaces    = false,
	breaklines          = true,
	backgroundcolor     = \color{white}
}
\begin{lstlisting}[language=Python]
outputs = model(inputs)
ce = criterion(outputs, targets)
loss = ce + 0.5*b*pow(outputs-c, 2).mean()       # Here it is!
optimizer.zero_grad()
loss.backward()
optimizer.step()
\end{lstlisting}

\section{Why Does Output Decay Help?}
\label{visualize}
In order to shed some light on the effect of Output Decay, we conducted a series of systematical experiments to seek potential reasons for better generalization.

\begin{figure}[h]
	\centering
	\includegraphics[width=0.97\columnwidth]{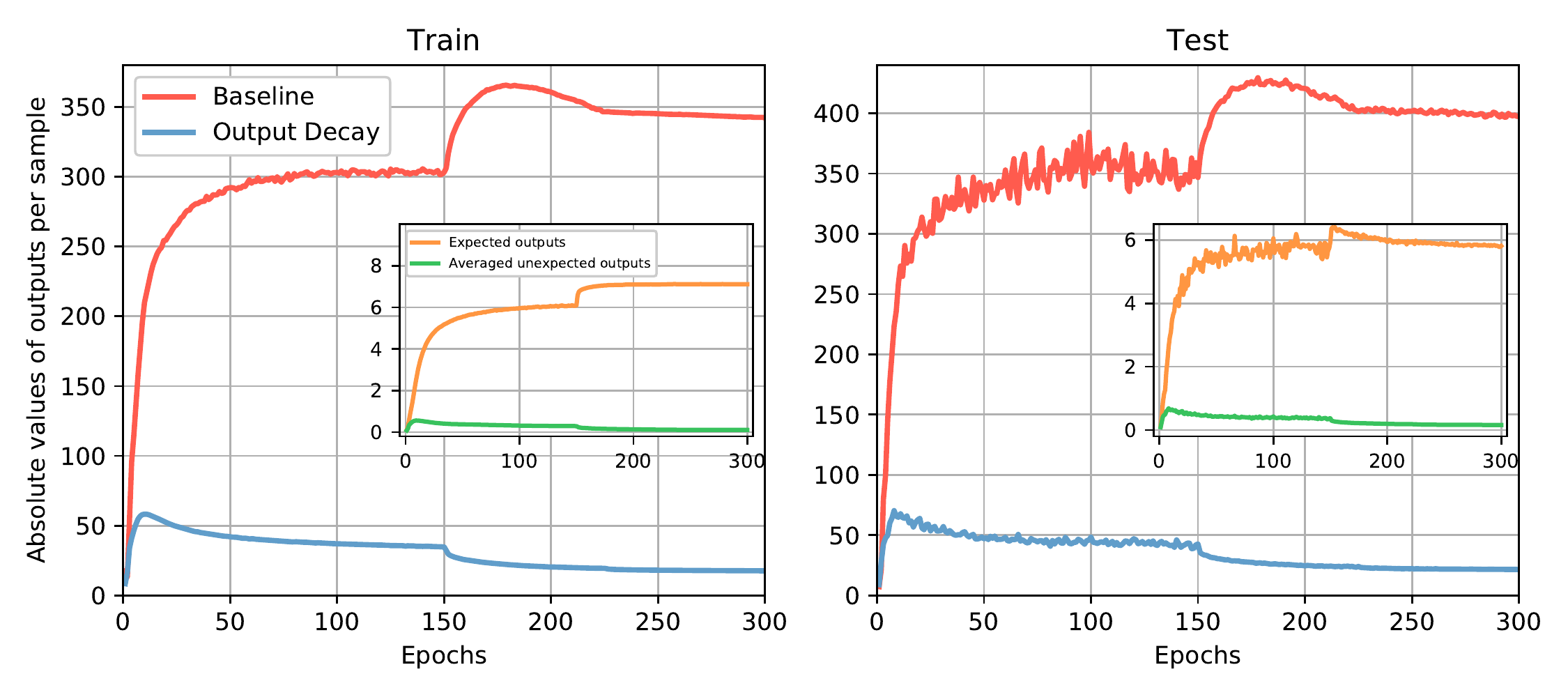}
	\caption{Changes of absolute values of outputs as training(left)/testing(right) proceeds. \textbf{Center Right Insets}: Comparison between absolute values of expected outputs and averaged unexpected outputs of OD. These results show while OD shrinks the magnitudes of outputs, the expected outputs can still get larger due to the original loss function.}
	\label{outputanalysis}
\end{figure}

\paragraph{Outputs}
We first explored whether the Output Decay could make the outputs smaller as training proceeds. To this end, we counted up the absolute values of outputs and averaged them over all samples. As seen in Figure~\ref{outputanalysis}, without exerting constraints on outputs, the outputs in baseline training became larger and larger during training. This trend stays the same while testing. Conversely, OD indeed shrinks outputs, as expected. 

We then delved deeper into the behavior into Output Decay. The insets in Figure~\ref{outputanalysis} depict the change of both expected (\ie outputs that correspond to the ground-truth label) and averaged unexpected (\ie sum of outputs that corresponds to other labels and divided by the number of those labels) outputs, produced by models trained with OD during training and testing. It can be easily observed that whereas the mean value of unexpected outputs takes on a decreasing change due to the regularized effect of OD, the expected outputs still manage to rise so that models can still maintain comparable representation abilities. These results are further proofs of the “compromising” behavior of OD mentioned earlier.

\paragraph{Calibration}
The above `compromising' phenomenon makes us wonder if OD can improve the calibration of the model by making the confidence of its predictions more accurately represent their accuracy. Calibrated confidence estimates are very important for model practicability, since a model with bad quality of confidence estimates is unreliable and can not be trusted with on some safety-critical applications such as autonomous driving or computer-aided diagnosis. To seek the answer to the question, we investigate the calibration of the model trained with and without OD.

\begin{figure}[h]
	\centering
	\includegraphics[width=0.8\columnwidth]{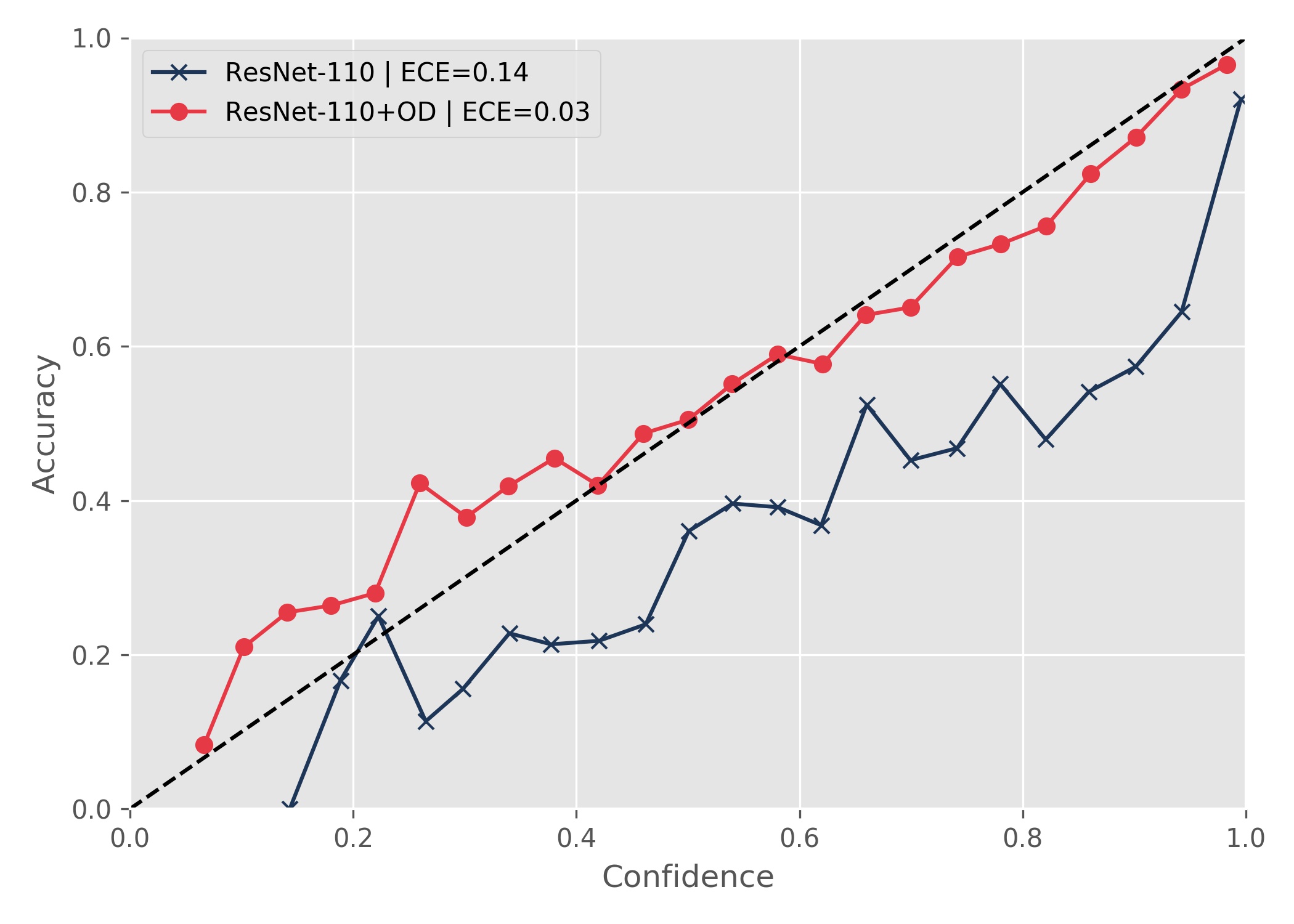}
	\caption{Reliability diagram for CIFAR-100 with ResNet-110 trained with and without OD. This graph demonstrates that OD helps produce well-ranked confidence estimates.}
	\label{confidence}
\end{figure}

Figure~\ref{confidence} depicts the 25-bin reliability diagrams of ResNet-110 trained with and without OD on CIFAR-100. We can see that the vanilla model is quite un-calibrated and tends to be overconfident since the accuracy is always lower than the confidence values. On the contrary, we then can observe the effects of OD on calibration that the slope of OD is very closer to the slope of 1. We also computed the expected calibration error (ECE)~\cite{naeini2015obtaining}. Again, OD achieves a better result. It is quite remarkable since such a simple method can lead to such well-ranked confidence estimates.

\begin{figure}[h]
	\centering
	\includegraphics[width=0.99\columnwidth]{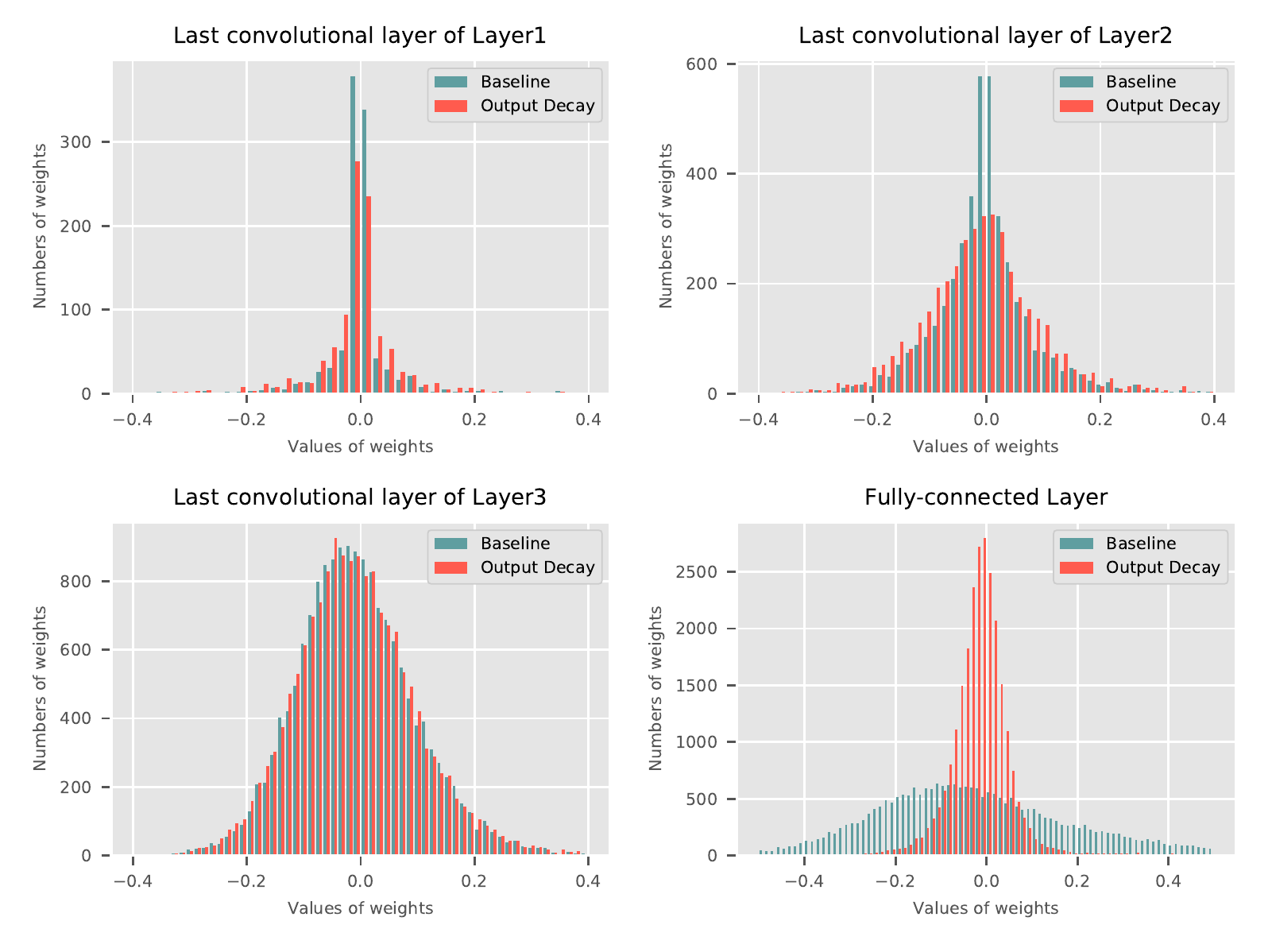}
	\caption{The distributions of parameters. A standard ResNet-110($Acc=74.04\%$) is compared with a ResNet-110 trained with OD($Acc=76.09\%$) at four different layers.}
	\label{parameter-ayalysis}
\end{figure}

\paragraph{Parameters}
Observations on changes of parameters are of vital importance in understanding the behavior of Output Decay. Figure~\ref{parameter-ayalysis} characterizes the distributions of parameters coming from a standard ResNet-110 and a ResNet-110 trained with OD. Interestingly, it appears that the major modification to parameters happens on the fully-connected layer, rather than the convolutional layers. By regularizing the output distributions, OD undercuts most large weights, which suggests that OD helps provide a simpler and more powerful explanation for the data. In addition, the smallness of weights brings the model an ability to resist underlying noise information of data. A network with smaller weights prefers to learn from evidence which is seen often across the training set while a network with large weights may change its behavior frequently in response to some specific information. Detailed empirical experiments of this effect will be shown in the following contents.

\begin{figure}[htbp]
	\centering
	\includegraphics[width=0.99\columnwidth]{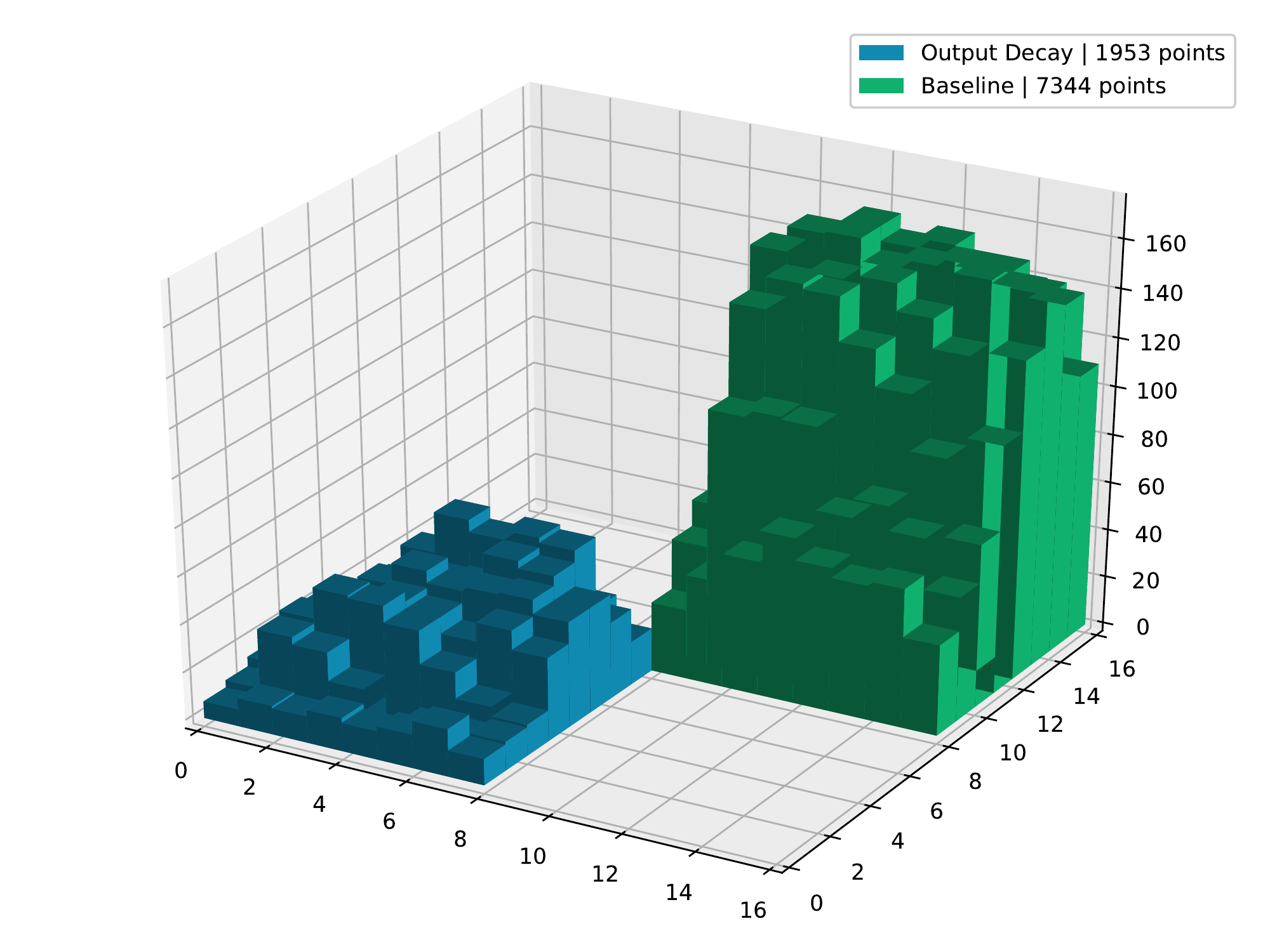}
	\caption{Visualizations of activation situation. The size of blue/green area is $8\times8$. The height($z$) of the bar represent the number of times the point at this position($x$-$y$) is activated. In a feature map with a total of 16384 points, OD activated 1953 points and baseline(without OD) activated 7344 points. The results present that the activation of network trained with OD is more sparse.}
	\label{activation-analysis}
\end{figure}

\paragraph{Activations of Feature Map}
To investigate how feature information flows through the network while applying Output Decay, we then visualize the feature activations after the last convolutional layer of the network. Specifically, we randomly selected a testing sample and extracted its corresponding feature map ($256\times8\times8$), denoted by $F^{C\times H\times W}$. For better illustration, we superimposed the statistics of 256 partial feature maps ($8\times8$) into a matrix $M^{H\times W}$. Since there were a large number of inactive points in the feature map, we chose to only count the number of points with values greater than 0. We can obtain that:
\begin{equation}
M_{i,j} = \sum_{c=1}^{C} \mathbbm{1}{(F[c, i, j]>0)}
\end{equation}
where $\mathbbm{1}$ denotes the indicator function and it tests a conditional expression. Then we plotted this matrix $M$ into a 3-D plane, where x-axis and y-axis denote $i$ and $j$, respectively. We used z-axis to denote the value of corresponding points, namely, each bar in Figure~\ref{activation-analysis} represents how many times these 256 partial feature maps have been activated at that position. We also provide the results obtained from the network trained with a regular scheme. 

As can be seen from the graph, while both demonstrating similar activation situations, there are fewer points being activated in the feature map from the last convolutional layer trained with OD. On one hand, this sparse activation situation symbolizes a stronger \textit{feature selection} ability, on the other hand, the sparse activation makes the network become more difficult to fit the data, implying that OD also \textit{suppresses the emergence of overfitting} to a certain extent.

\begin{figure}[htbp]
	\centering
	\includegraphics[width=0.99\columnwidth]{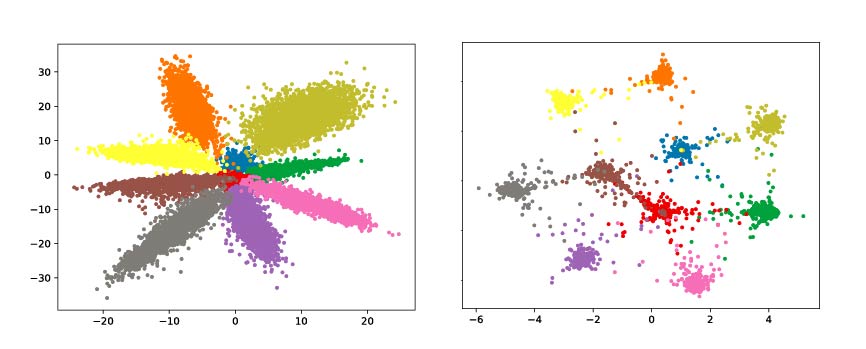}
	\caption{Visualizations of 2-D feature distribution on ResNet-110 on CIFAR10. Left plot: Trained without OD; Right plot: Trained with OD. Results show that OD helps obtain more discriminative features.}
	\label{feature-distribution}
\end{figure}

\paragraph{Visualizations of Feature Distributions}
In order to intuitively understand the effect of Output Decay on the network, we perform a visualization scheme similar to~\cite{wen2016discriminative}. It is based on three steps: (1) Add a new fully-connected layer between the last convolutional layer and the original fully-connected layer. The output number of this layer is 2 so that we can directly plot its features onto a plane. (2) Fine-tune the input number of the original fully-connect layer. (3) Train and project all the outputs of this new layer onto a 2-D plane. Figure~\ref{feature-distribution} compares the visualization results obtained from the networks trained with and without OD.

This graph is quite revealing in several ways: (1) under the supervision of cross-entropy loss, the deeply learned features are separable, but (2) they are not discriminative enough, as evidenced by the significant intra-class variations. By contrast, (3) the projections of OD are spread into 10 defined and tight clusters, demonstrating that OD results in better discriminative feature learning in terms of smaller intra-class variations and larger inter-class distances.

\section{Experiments}
\label{experiments}
\subsection{Output Decay can be universally applied to various architectures.}
We first extensively evaluate Output Decay on various architectures: Vgg~\cite{simonyan2014very}, MobileNets~\cite{howard2017mobilenets, sandler2018mobilenetv2}, ResNet~\cite{he2016deep}, preResNet~\cite{he2016identity}, ResNext, WideResNet~\cite{zagoruyko2016wide}, DenseNet~\cite{huang2017densely}, and PyramidNet~\cite{han2017deep}. All the experiments were performed on CIFAR-100~\cite{krizhevsky2009learning} that contains 50000 training and 10000 testing images divided into 100 classes. For fair comparison, all the experiments strictly use the same preprocessing and data augmentation strategies such as random flipping and cropping, following the common practice. 
All the hyper-parameters and architectural details remain unchanged too (See Appendix~\ref{cifar100detail} for full implementation details). All experiments were conducted four times and the averaged best performances during training are reported.

Notably, the results in Table~\ref{cifar-results} show that OD consistently offers significant performance gains over the baselines, such as an impressive 1.37$\%$ top-1 error reduction on ResNet-110. We would like to note the \textit{practical and theoretical implications} provided by Output Decay. Practically, any CNN-based architecture can acquire performance gains through negligible adjustments. Theoretically, the fact that such a small alteration results in such a significant improvement could offer much food for thought regarding the nature and extent of output regularization. 

\begin{table}[h]
	\begin{center}
		\begin{small}
			\begin{tabular}{p{4cm}|p{1.46cm}p{1.46cm}}
				\toprule[1.3pt]
				\midrule
				Model & Vanilla & + Ours\\
				\midrule
				Vgg19$\_$bn		  &28.45{\scriptsize $\pm$0.11}   & \textbf{27.82}{\scriptsize $\pm$0.05}\\
				\midrule
				MobileNet           &31.54{\scriptsize $\pm$0.17}  & \textbf{30.27}{\scriptsize $\pm$0.15}\\
				MobileNetV2         &27.33{\scriptsize $\pm$0.12}  & \textbf{26.81}{\scriptsize $\pm$0.09}\\
				\midrule
				ResNet-56         &26.34{\scriptsize $\pm$0.01}   & \textbf{25.08}{\scriptsize $\pm$0.13} \\
				ResNet-110        &25.47{\scriptsize $\pm$0.09}   & \textbf{24.1}{\scriptsize $\pm$0.02}\\
				\midrule
				ResNet-56-PreAct  &25.82{\scriptsize $\pm$0.06}   & \textbf{25.07}{\scriptsize $\pm$0.01}\\
				\midrule
				DenseNet-100-12   &22.78{\scriptsize $\pm$0.04}   &\textbf{21.72}{\scriptsize $\pm$0.03}\\
				\midrule
				WRN-28-10         &18.79{\scriptsize $\pm$0.05}  & \textbf{18.43}{\scriptsize $\pm$0.06}\\
				\midrule
				ResNext-8-64         &18.09{\scriptsize $\pm$0.17}  & \textbf{17.51}{\scriptsize $\pm$0.06}\\
				\midrule
				PyramidNet-110-270\tiny(BasicBlock)  &18.63{\scriptsize $\pm$0.10}  & \textbf{17.92}{\scriptsize $\pm$0.08}\\
				PyramidNet-200-240\tiny(Bottleneck)  &17.09{\scriptsize $\pm$0.13}  & \textbf{16.61}{\scriptsize $\pm$0.16}\\
				\bottomrule[1.3pt]
			\end{tabular}
		\end{small}
	\end{center}	
	\caption{Top-1 Error($\%$) with and without OD using different architectures on CIFAR-100.}
	\label{cifar-results}
\end{table}

\subsection{Output Decay is equally effective on multiple benchmark datasets.}
We then extended the experiments to other benchmark datasets. In particular, we used the following five benchmark datasets: Fashion-MNIST~\cite{xiao2017/online}, SVHN~\cite{netzer2011reading}, CIFAR-10~\cite{krizhevsky2009learning}, Tiny-ImageNet, ImageNet~\cite{russakovsky2015imagenet}. Together with the previous CIFAR-100, we believe these six datasets should be able to cover a wide range of scenarios in the computer vision field. The implementation details of these datasets can be found in Appendix~\ref{datasetdetail}.

We show the results in Table~\ref{dataset-results}. These results prove the wide applicability on various datasets. We draw attention to the fact that OD performs even better on larger and harder datasets, which is not surprising as the potential of models trained on small datasets has nearly been drained out. 

\begin{table}[h]
	\begin{center}
		\begin{small}
			\begin{tabular}{llll}
				\toprule[1.3pt]
				\midrule
				Datasets& Model & Vanilla & + Ours\\
				\midrule
				Fashion-MNIST    & ResNet-29		  &4.83{\scriptsize $\pm$0.02}   & \textbf{4.64}{\scriptsize $\pm$0.01}\\
				\midrule
				SVHN             & ResNet-29		  &4.3{\scriptsize $\pm$0.03}   & \textbf{3.82}{\scriptsize $\pm$0.02}\\
				\midrule
				CIFAR-10         & ResNet-29		  &7.16{\scriptsize $\pm$0.06}   & \textbf{6.68}{\scriptsize $\pm$0.06}\\
				\midrule
				Tiny-ImageNet   & ResNet-110	      &37.62{\scriptsize $\pm$0.04}   & \textbf{36.04}{\scriptsize $\pm$0.11}\\
				\midrule
				ImageNet         & ResNet-50		  &23.77{\scriptsize $\pm$0.02}   & \textbf{22.68}{\scriptsize $\pm$0.01}\\
				\bottomrule[1.3pt]
			\end{tabular}
		\end{small}
	\end{center}	
	\caption{Top-1 Error($\%$) with and without OD on multiple benchmark datasets.}
	\label{dataset-results}
\end{table}

\begin{figure}[h]
	\centering
	\includegraphics[width=0.99\columnwidth]{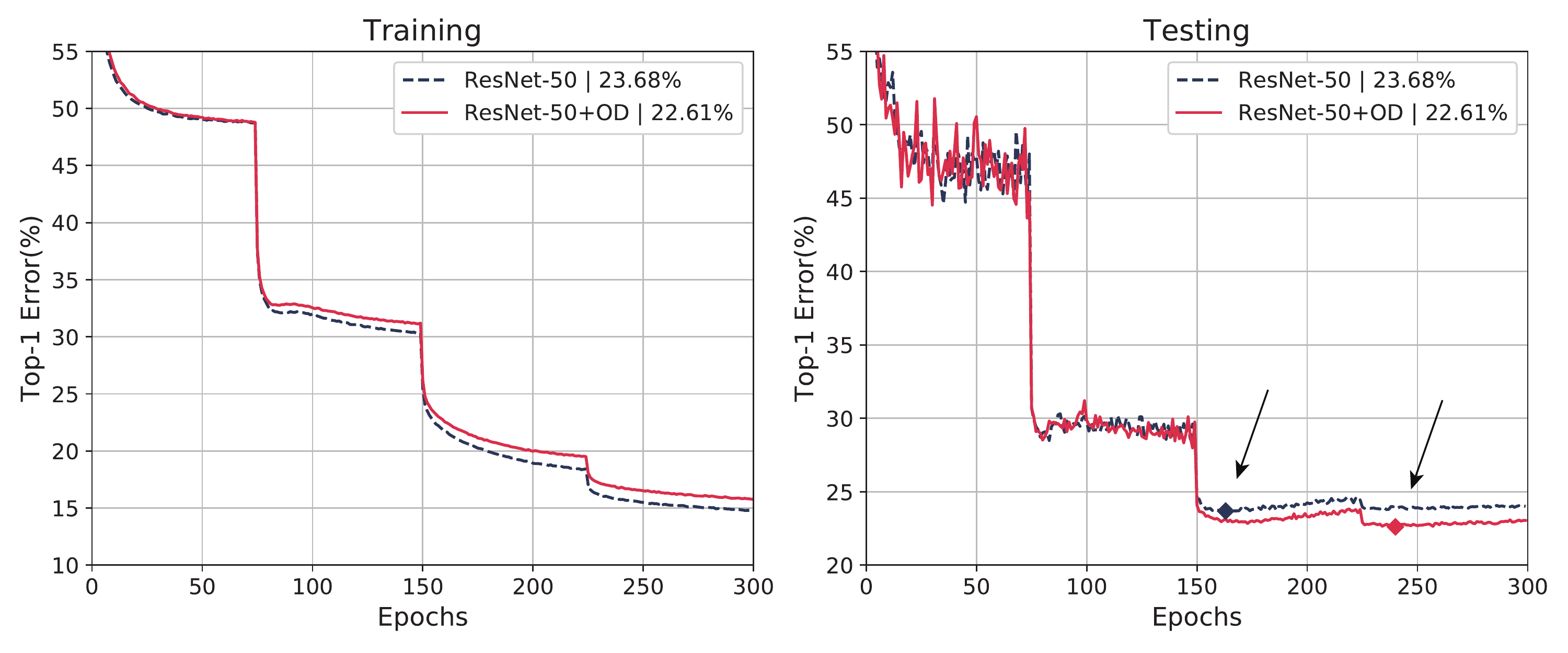}
	\caption{Top-1 training and testing error plot on ImageNet Classification. Left plot: training; right plot: testing. The arrows point to the timings where two models achieve their best performances. OD helps acquire further increase even in the late stage of training.}
	\label{trainingcurve}
\end{figure}

We then inspect the training and testing curves (in terms of top-1 error w.r.t epoch numbers) of models with and without OD on ImageNet. As shown in Figure~\ref{trainingcurve}, the vanilla ResNet-50 model achieves smaller training errors than OD throughout the whole training. But OD significantly outperforms the baseline on the testing data, implying that OD prevents the model from fitting the data too well for better generalization. As a byproduct, OD produces a double decent curve and achieves further improvements after epoch 225, when the vanilla model have suffered from overfitting due to the change of learning rate.

\subsection{Output Decay is compatible with various techniques.}
In practice, scenarios in which more than one regularization or data augmentation techniques are used occur frequently. Therefore, compatibility with other techniques is also an important indicator of practical implication. Hence, we further apply Output Decay with some existing commonly used techniques, such as Cutout~\cite{devries2017improved}, Mixup~\cite{zhang2018mixup}, Autoaugment~\cite{cubuk2018autoaugment}, FashAutoAugment~\cite{lim2019fast}, Label Smooth~\cite{szegedy2016rethinking}, and Confidence Penalty~\cite{pereyra2017regularizing}. To ensure the effectiveness of the proposed method, a strong baseline is used here, which is PyramidNet-110 with widening factor $\alpha = 270 $. The detailed implementation settings are described in Appendix~\ref{combinationdetail}.

\begin{table}[h]
	\begin{center}
		\begin{small}
			\begin{tabular}{llll}
				\toprule[1.3pt]
				\midrule
				Model & Method &  Top-1  & Top-5  \\
				\midrule
				\multirow{12}{*}{PyramidNet}& +Cutout           & 17.52 & 4.01 \\
				& +Cutout+OD        & \textbf{16.94} & 4.20  \\ 
				& +Mixup            & 16.90 & 4.35 \\
				& +Mixup+OD         & \textbf{16.53} & 4.18   \\
				
				\cmidrule(r){2-4}
				& +AutoAugment               & 17.20 & 3.68 \\
				& +AutoAugment+OD            & \textbf{16.16} & 3.72 \\
				& +FastAutoAugment           & 17.43 & 3.60 \\
				& +FastAutoAugment+OD        & \textbf{16.66} & 3.83 \\
				\cmidrule(r){2-4}
				& +LabelSmooth                 & 18.56 & 5.13 \\
				& +LabelSmooth+OD              & \textbf{18.22} & 5.05 \\
				& +ConfidencePenalty           & 18.41 & 4.92 \\
				& +ConfidencePenalty+OD        & \textbf{17.90} & 4.68 \\
				\bottomrule[1.3pt]
			\end{tabular}
		\end{small}
	\end{center}	
	\caption{Top-1 and Top-5 Error($\%$) of PyramidNet-110-270(BasicBlock, the top-1 error of the vanilla model is 18.63$\%$) with the combination of OD and some commonly used techniques on CIFAR-100. Results demonstrate the wide compatibility of OD.}
	\label{combination}
\end{table}

It can be seen from the data in Table~\ref{combination} that OD is able to make further improvements, even on some strong techniques. We want to note again that though these improvements are not as high as one might expect, the improvements due to OD come at zero architectural alteration, nearly one line of code modification, negligible memory use and computational time increase.

\subsection{Output Decay can still work in some practical scenarios. }
\label{practical}
\noindent\textbf{A. Label Noise Problem}  The training of the model is closely related to the quality of the data. However, due to the expensive and time-consuming labeling process, real-world datasets~\cite{learning2015xiao, cleannet2018lee} always have label noise problems, which makes models hard to achieve desirable performances. Motivated by the above deduction that models with smaller weights tend to prioritize learning patterns rather than special cases and thereby would have more label noise robustness, we set up experiments with noisy datasets to see how well Output Decay performs for different types and amounts of label noise.

Following~\cite{coteaching2018han, distilling2019zhang, probabilistic2019yi}, we corrupted these datasets manually on CIFAR-10 and CIFAR-100. Two types of noisy label are considered: (1) Symmetry flipping: each label is set to an incorrect value uniformly with probability; (2)Pair flipping: labelers may only make mistakes within very similar classes. We use $\epsilon$ to denote the noise rate. We provide illustrations of these two types in Appendix~\ref{noisedetail}

Table~\ref{noisy} compares the results with and without OD using ResNet-56. Regardless of different types and amounts of noise, OD has all made huge improvements on baseline (which only uses Cross Entropy to classify). We also draw attention to the fact that OD performs better on pair flipping noise, which is known as a more realistic setting since it corrupts semantically-similar classes. One possible implication of this is that it is an unexpected side product of OD when increasing inter-class distances. Together these results provide strong proofs that OD helps raise the internal noise tolerance of the classifier. 

\begin{table*}[h]
	\begin{center}
		\begin{small}
			\begin{tabular}{lcccc}
				\toprule[1.3pt]
				\midrule
				\multirow{2}{*}{Noisy CIFAR-10}  & \multicolumn{2}{c}{Symmetric Flipping} & \multicolumn{2}{c}{Pair Flipping} \\
				\cmidrule(r){2-3}\cmidrule(r){4-5}
				& $\epsilon=0.5$            & $\epsilon=0.2$           & $\epsilon=0.45$ &    $\epsilon=0.2$\\
				\midrule
				Baseline                         & 78.53             & 88.74            & 73.59   & 89.86   \\ 
				Baseline+OD                      & \textbf{81.74} ($+ 4.09\% $)            & \textbf{89.62} ($+ 0.99\% $)      & \textbf{85.16} ($+ 15.72\% $)& \textbf{90.44} ($+ 0.65\% $) \\ 
				\midrule[1.3pt]
				\multirow{2}{*}{Noisy CIFAR-100} & \multicolumn{2}{c}{Symmetric} & \multicolumn{2}{c}{Pair} \\
				\cmidrule(r){2-3}\cmidrule(r){4-5}
				& $\epsilon=0.5$            & $\epsilon=0.2$           & $\epsilon=0.45$ & $\epsilon=0.2$ \\
				\midrule
				Baseline                         & 45.55                         & 62.67            & 41.90   & 64.43\\ 
				Baseline+OD                      & \textbf{57.37} ($+ 25.95\% $)          & \textbf{69.03} ($+ 10.16\% $)            & \textbf{51.77} ($+ 23.56\% $)    &\textbf{71.88} ($+ 11.56\% $)\\
				\bottomrule[1.3pt]
			\end{tabular}
		\end{small}
	\end{center}
	\caption{Average test accuracy on noisy CIFAR-10 and CIFAR-100 using ResNet-56. The best accuracy is indicated as bold, and we use brackets to report the relative  gains over each counterpart without OD. Results show that OD helps achieve label noise robustness.}
	\label{noisy}
\end{table*}

\noindent\textbf{B. Small Sample Size Problem} Most benchmark datasets we are using now have sufficient training samples for each class. However, in some real industrial scenarios, the size of dataset may be small, which means that there are not enough training samples for each class. Consequently, this is a huge challenge for the feature selection ability of the network. Moreover, according to~\cite{zhang2016understanding, arpit2017closer}, over-parameterized deep networks can gradually memorize the data, and fit everything in the end. Hence, the small sample size also means that the network may quickly overfit.

\begin{figure}[h]
	\centering
	\includegraphics[width=0.99\columnwidth]{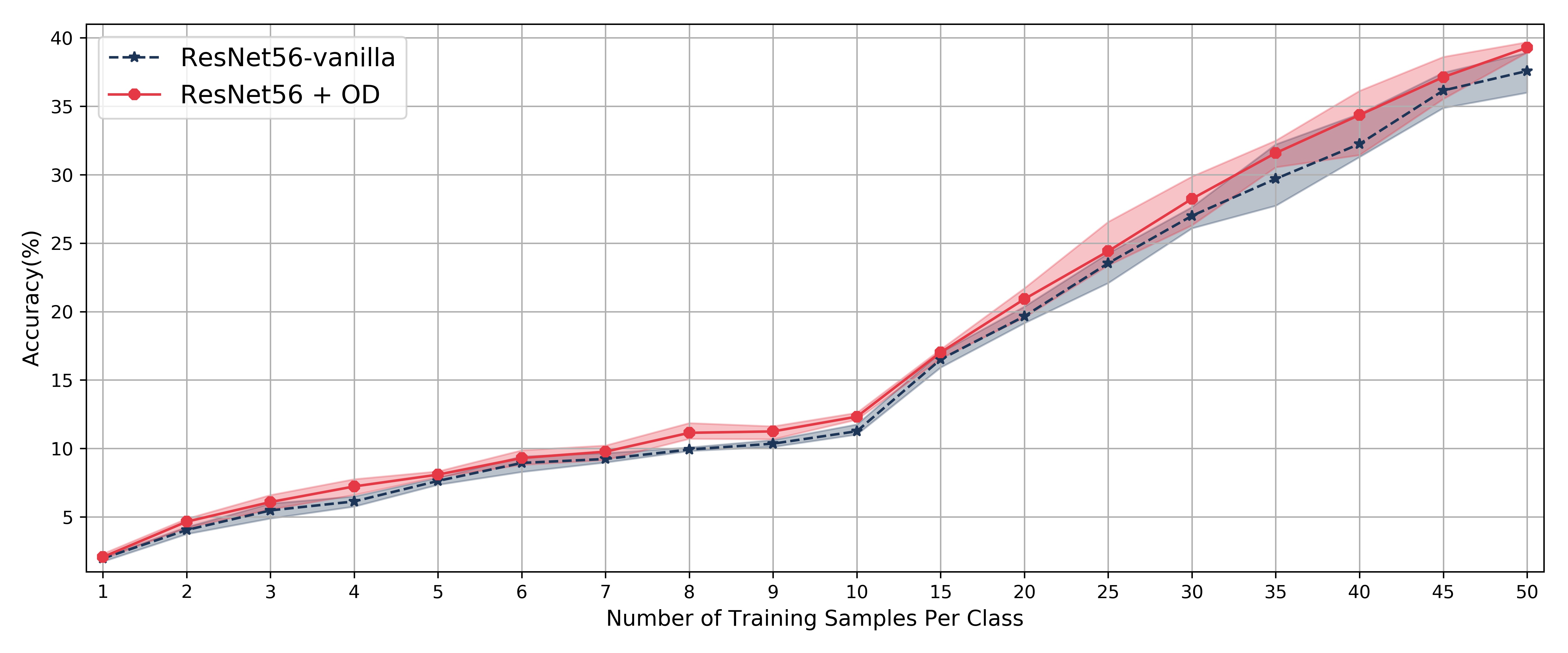}
	\caption{Accuracy($\%$) of ResNet-56 with and without OD under varying number of training samples per class in CIFAR-100. The shaded areas denote the minimum and maximum results from four runs. OD achieves consistent improvements over the vanilla ResNet-56.}
	\label{smalldata}
\end{figure}

To evaluate the effectiveness of OD, we constructed a series of sub-datasets of CIFAR-100 via randomly choosing $n\in \{1,2,3,...9,10,15, 20,...,45,50\} $ samples for each class. Figure~\ref{smalldata} compares the results obtained from ResNet-56 trained with and without OD. It’s worthy of noting that we remain the hyper-parameter setting unchanged since the model has not changed. There might be a chance that the optimal setting of OD for the model will change as the size of the dataset changes. Future studies on this topic are therefore recommended.

From the graph, we can see that OD consistently outperforms the vanilla setting on all of the sub-datasets, in accordance with the expectations. This finding broadly supports the previous speculations on OD that: (1) the sparse feature activation does represent a stronger feature selection capability of the network trained with OD; (2) OD can cope with overfitting to some extent.

\subsection{Ablation Studies}
\begin{figure}[h]
	\centering
	\includegraphics[width=0.99\columnwidth]{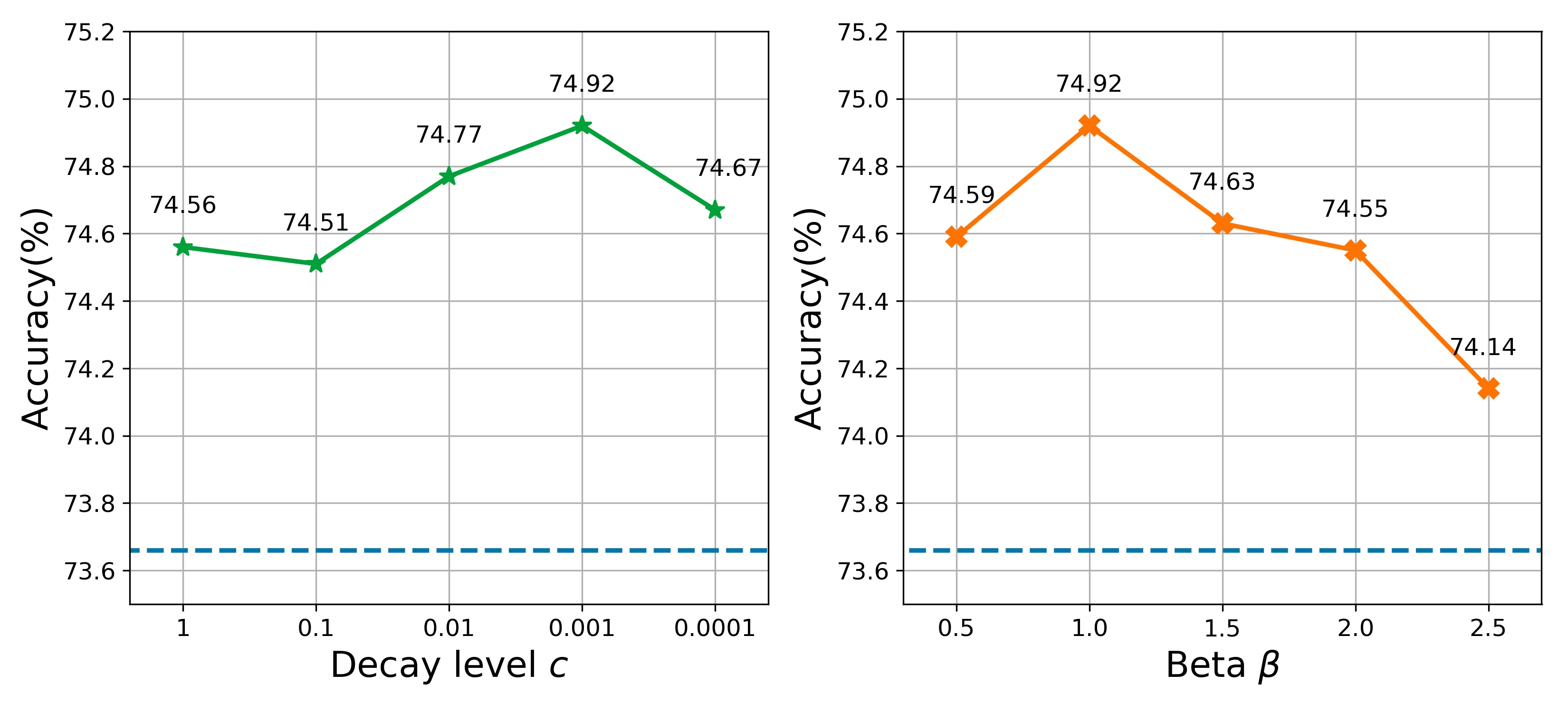}
	\caption{Effect of the decay level $c$ and coefficient $\beta$ on CIFAR-100 classification accuracy using ResNet56. The blue dashed lines denote the baseline accuracy.}
	\label{parameter}
\end{figure}

Here we conducted ablation studies on CIFAR-100 dataset using ResNet-56.
\paragraph{Output Decay is robust to parameter changes to a large extent.} We first explore the effect of hyper-parameters. Two parameters need to be evaluated, the decay level $c$ and the coefficient $\beta$. As for the decay level $c$, it denotes the value we expect each component of outputs to achieve. To see the impact of this parameter in isolation, we set $\beta=1$ and evaluate the network with different $c \in \{10^0, 10^{-1}, 10^{-2}, 10^{-3}, 10^{-4}\}$. The results are presented in Figure~\ref{parameter}, left plot. It can be seen that when the decay level decrease, the performance of the network first increases, reaches the top, and then decreases. This result may be explained by the fact that under a very small decay level, we are in fact imposing a very large penalty, which can make it difficult to learn at the beginning and thereby lead to performance degradation. But it is worthy of noting that OD outperforms the baseline consistently for all of the choices of $c$.

Regarding the coefficient $\beta$, almost all the regularization methods would have such a similar parameter, to indicate whose choice should play an important role in the regularized training process. When evaluating $\beta$, we remain $c=10^{-3}$ unchanged, and adjust $\beta$ from $\{0.5, 1, 1.5, 2, 2.5\}$. From the right plot of Figure~\ref{parameter} we find OD rather insensitive to the choice of $\beta$. Again, the accuracy of OD is higher than baseline for all $\beta$ considered, demonstrating that OD is robust to parameter changes to a large extent.

\paragraph{Different choices of OD can also bring some improvement.}
A careful examination of every detail of the method is necessary. To this end, we try several design possibilities of OD and classify these choices into three groups. 

The first group focuses on exploring different loss functions. OD-KL uses the KL loss function to make the output distributions of the network approximate to a uniform distribution. OD-L1 replaces the Mean Squared Error loss used in the original OD with L1 loss function. The second group tests whether there are better design choices of coefficient $\beta$. OD-LinearUp linearly increases the coefficient $\beta$ as training proceeds while OD-LinearDown linearly decreases $\beta$ to reduce the weight of OD. As the strategy of WarmUp has proved useful in many practical uses, OD-WarmUp uses it to modify $\beta$ so that $\beta$ rises from 0 to 1 in the first 75 rounds and then slowly drops to 0. The last group explores a different choice of the decay level $c$. Since the original OD gives a final direction at the beginning, we tried to see if we could reduce the outputs dynamically, thus giving birth to the OD-Dynamic. When applying OD-Dynamic, in each round we calculate the average of all the outputs of the current round and use the same form to make the output of the network approximate this value. It means that the decay level varies with each round.

Table~\ref{variants} compares the results of different design choices of OD. One interesting note is that all the variants of OD improve upon the baseline to some extent, representing that the idea and the direction of regularizing outputs is correct and forward-looking. However, all the variants also result in certain performance degradation to some extent. This suggests that these variants require further exploration of parameter search. In view of a comprehensive consideration of interpretability, simplicity and performance, we stick to the original scheme of Output Decay.

\begin{table}[h]
	\begin{center}
		\begin{small}
			\begin{tabular}{p{5cm}p{2cm}}
				\toprule[1.3pt]
				\midrule
				Model(ResNet-56) & Top-1 Error($\%$)\\
				\midrule
				Baseline              & 26.34{\scriptsize $\pm$0.01}   \\
				Proposed(OD)         & \textbf{25.08}{\scriptsize $\pm$0.13}\\
				\midrule
				OD-KL 				&26.34{\scriptsize $\pm$0.03} \\
				OD-L1				&25.81{\scriptsize $\pm$0.05}\\
				\midrule
				OD-LinearUp            &25.81{\scriptsize $\pm$0.01}\\
				OD-LinearDown       &25.47{\scriptsize $\pm$0.15}\\
				OD-WarmUp           &25.80{\scriptsize $\pm$0.02}\\
				\midrule
				OD-Dynamic         &25.70{\scriptsize $\pm$0.05}\\
				\bottomrule[1.3pt]
			\end{tabular}
		\end{small}
	\end{center}	
	\caption{Performance of different design choices of Output Decay with ResNet-56 on CIFAR-100.}
	\label{variants}
\end{table}

\section{Related Works}
\noindent\textbf{Data augmentation methods.} Focusing on adding randomness and uncertainty on the input images, data augmentation methods~\cite{yun2019cutmix, devries2017improved, cubuk2018autoaugment, lim2019fast, zhong2020random} have long been used frequently in computer vision. Aiming in a different direction, OD not only is compatible with these methods but also can make further improvements while combining with them.

\noindent\textbf{Label smoothing and confidence penalty.} The critical difference between label smoothing~\cite{szegedy2016rethinking} and OD is that label smoothing operates on labels while OD operates on outputs. Moreover, label smoothing mainly penalizes the ground-truth labels while OD enforces all the outputs to be as small as possible. Regarding confidence penalty~\cite{pereyra2017regularizing}, OD shares certain similarity with confidence penalty that they both focus on regularizing the output distribution. However, confidence penalty is merely penalizing the negative entropy to prevent peaked distributions while OD wants the output to be smaller and more uniformly distributed. These three methods are therefore different from each other. More importantly, results have empirically demonstrated that OD outperforms the other two and can be combined with them.

\noindent\textbf{Temperature scaling. }Temperature scaling~\cite{guo2017on} is an effective method on calibration method, but it does not affect the model’s accuracy while OD brings better classification performance. Not only that, temperature scaling shrinking the outputs using the same scaling while OD imposes different penalties on each component of outputs.

\noindent\textbf{Weight decay. }While both weight decay~\cite{krogh1991a} and OD can shrinking weights of networks, the form of weight decay and OD are completely different, which means the gradients and the learning rule brought by these two methods are completely different. The most important thing is that all the experiments are performed using weight decay including each baseline and its OD counterpart. These suggest that OD is a regularization method with the ability to further improve the performance on the basis of weight decay.

\section{Discussion}
We have introduced Output Decay that regularizes outputs to improve the performance and generalization of networks. It’s simple that it can be conveniently incorporated with CNNs without any other change entailed. Experiments empirically demonstrated that OD can be applied to various architectures, datasets, and classification scenarios, including confidence estimates, label noise, and small data regime. 

Despite its exploratory nature, this study offers some insight into how to operate on outputs and why brings improvements. It is hoped that this work may generate fresh insights into deep learning.

{\small
	\bibliographystyle{ieee_fullname}
	\bibliography{egbib}
}

\clearpage
\appendix
\section{More Observations on ResNet and DenseNet Architectures.}
\label{more_observation}
We have already mentioned the observations on Vgg architectures in the main text of the article. Here we present what we observed on another two architectures (ResNet and DenseNet). All of these models are also the pretrained publicly available models provided by \textit{Pytorch}.

\begin{figure}[h]
	\centering
	\includegraphics[width=0.99\columnwidth]{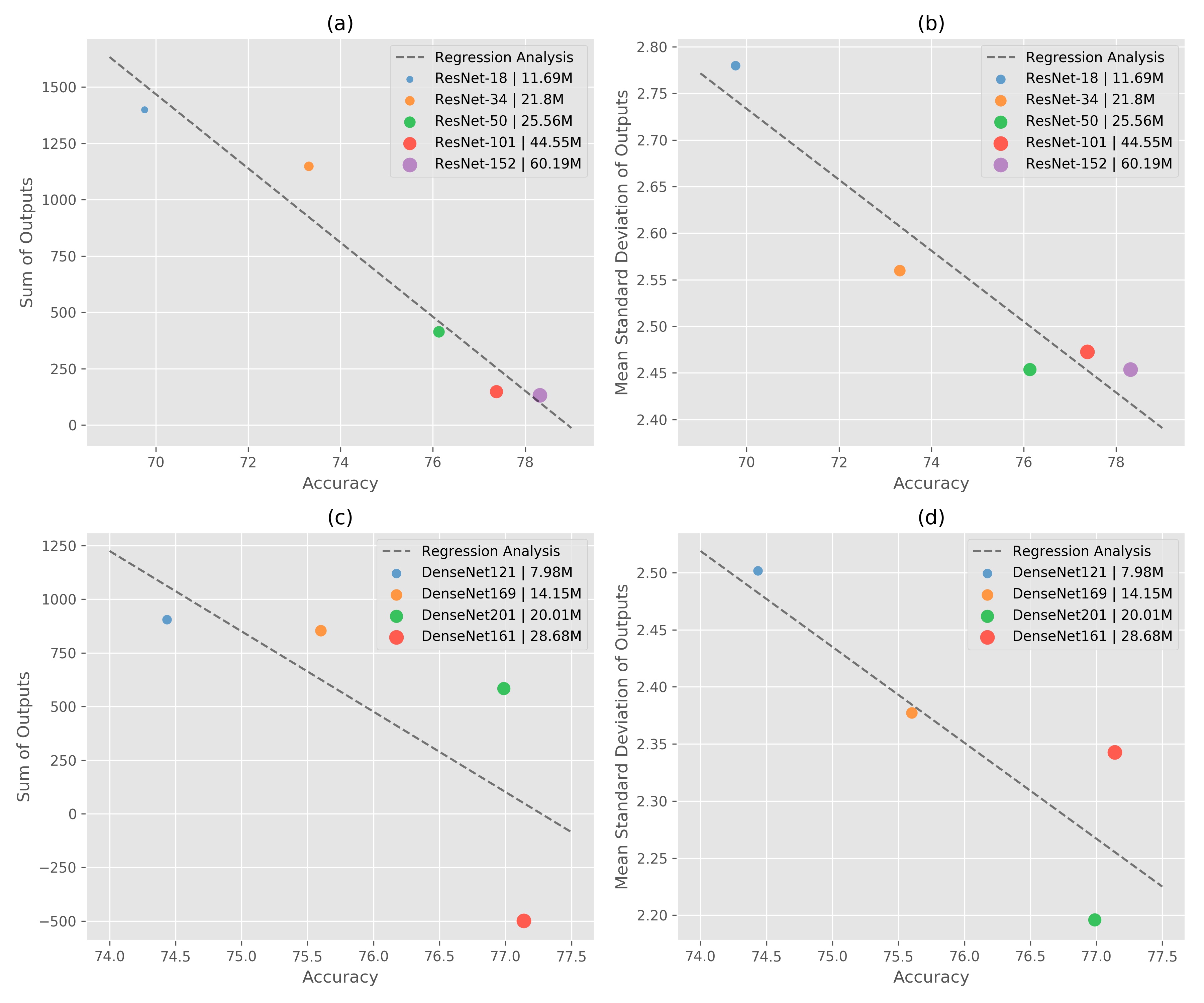}
	\caption{Connections between the models' performance and magnitudes(\textbf{a}, \textbf{c})/variations(\textbf{b}, \textbf{d}) of the model's outputs on ResNet(\textbf{a}, \textbf{b}) and DenseNet(\textbf{c}, \textbf{d}). Node size represents size of parameters. The dashed lines denote the estimated results of regression analysis. }
	\label{motivation-curve}
\end{figure}
We can clearly observe similar patterns that the accuracy of models is negatively correlated with the mean values (For ResNets: $R^2 = 0.944, *p=0.006<0.05$; For DenseNets:$R^2 = 0.555, p=0.255<0.5$) and mean variances (For ResNets: $R^2 = 0.905, *p=0.013<0.05$; For DenseNets:$R^2 = 0.698, p=0.164<0.5$) of the output distributions.

\section{Implementation Details on CIFAR-100.}
\label{cifar100detail}
We used a total of eight different network architectures to demonstrate the wide applicability and compatibility of OD. For ResNet~\cite{he2016deep}, pre-activation ResNet~\cite{he2016identity}, and Vgg~\cite{simonyan2014very}, we set mini-batch size to 128 and training epochs to 300. The initial learning rate was set to 0.1 and decayed by the factor of 0.1 at 150 and 225 epoch, respectively. We used a weight decay of 1e-4. For ResNext~\cite{xie2017aggregated}, we changed the weight decay to 5e-4. For MobileNets~\cite{howard2017mobilenets, sandler2018mobilenetv2}, since there weren't clear descriptions stated in papers, we used the same hyperparameter setting as used in ResNet. We changed the batch size to 64 when training with DenseNet~\cite{huang2017densely}. For Wide Residual Networks~\cite{zagoruyko2016wide}, we trained for 200 epochs with a weight decay of 5e-4. The learning rate was initially set to 0.1 but decreased by 5 at 60, 120 and 160 epochs. As mentioned in~\cite{han2017deep}, we changed the initial learning rate to 0.5 and decreased it by a factor of 0.1 at 150 and 225 epochs when training PyramidNet.

\section{Implementation Details on Different Datasets.}
\label{datasetdetail}
We have introduced the wide applicability of OD on various benchmark datasets. Here we describe the implementation details on these datasets. For Fashion-MNIST, SVHN, CIFAR-10, and Tiny-ImageNet, we adopted the same network training hyper-parameter setting as ResNet in CIFAR-100 (\ie mini-batch size 128, 300 epochs, learning rate 0.1, weight decay 1e-4). For ImageNet, we stayed the initial learning rate unchanged but reduced it by 10 at 75, 150, and 225 epochs, as done by \cite{yun2019cutmix}.

Regarding the details of augmentation strategies, different datasets require different normalizing parameters. For SVHN, we didn't adopt other strategies except normalization. For Fashion-MNIST, the input image was first randomly cropped to 28$\times$28 and then horizontally flipped. For CIFAR-10, the input image was first randomly cropped to 28$\times$28 and then horizontally flipped. For ImageNet, except for these two standard augmentation techniques, we also adopted colorjittering and lighting, as done in~\cite{yun2019cutmix}.

\section{Implementation Details when combined with other techniques.}
\label{combinationdetail}
Here we briefly discuss the settings for the techniques used in the main text.
When using Cutout, we selected the hole size of 8$\times$8 based on the result reported on~\cite{devries2017improved}. When using Mixup, we set $\alpha=\beta=0.5$ as suggested by~\cite{zhang2018mixup}. The second set of techniques is on the theme of data augmentation methods. For both AutoAugment~\cite{cubuk2018autoaugment} and FastAutoAugment~\cite{lim2019fast}, we modified the implementations based on the publicly available code.\footnote{\tiny{AutoAugment:\url{https://github.com/DeepVoltaire/AutoAugment/blob/master/autoaugment.py}\\FastAutoAugment: \url{https://github.com/ildoonet/cutmix/blob/master/autoaug/archive.py}}} Turing now to the final set with the topic of regularization, we chose Label Smoothing and Confidence Penalty. For Label Smoothing, we set the smoothing parameter $\epsilon=0.1$ as suggested by~\cite{szegedy2016rethinking}. For confidence penalty, we used a a confidence penalty weight of 0.1, following~\cite{pereyra2017regularizing}.

\section{More Explanations about Label Noise Problem}
\label{noisedetail}
Experiments on two settings about label noise are conducted:(1) Symmetric flipping: the label is independently changed to a random class with probability $\epsilon$. (2) Pair flipping: the label would be only flipped to a similar class with probability $\epsilon$. Consider a noise transition matrix $T$, where $T=Pr(\tilde{y}=j | y=i)$. $\tilde{y}$ is the flipper from $y$. Then the precise definitions of these two settings are as follow. We use 5 classes as an example.

\begin{equation*}
T_{\footnotesize{symmetric}}=\begin{bmatrix}
1-\epsilon & \frac{\epsilon}{4} & \frac{\epsilon}{4} & \frac{\epsilon}{4} & \frac{\epsilon}{4} \\
\frac{\epsilon}{4} & 1-\epsilon & \frac{\epsilon}{4} & \frac{\epsilon}{4} & \frac{\epsilon}{4} \\
\frac{\epsilon}{4} & \frac{\epsilon}{4} & 1-\epsilon & \frac{\epsilon}{4} & \frac{\epsilon}{4} \\
\frac{\epsilon}{4} & \frac{\epsilon}{4} & \frac{\epsilon}{4} & 1-\epsilon & \frac{\epsilon}{4} \\
\frac{\epsilon}{4} & \frac{\epsilon}{4} & \frac{\epsilon}{4} & \frac{\epsilon}{4} & 1-\epsilon \\
\end{bmatrix}
\end{equation*}

\begin{equation*}
T_{pair} = \begin{bmatrix}
1-\epsilon & \epsilon & 0 & 0 & 0 \\
0 & 1-\epsilon & \epsilon & 0 & 0 \\
0 & 0 & 1-\epsilon & \epsilon & 0 \\
0 & 0 & 0 & 1-\epsilon & \epsilon \\
\epsilon & 0 & 0 & 0 & 1-\epsilon \\
\end{bmatrix}
\end{equation*}

\end{document}